\documentclass[lettersize,journal]{IEEEtran}

\usepackage{amsmath,amsfonts}
\usepackage{algorithm}
\usepackage{algpseudocode}
\usepackage{array}
\usepackage[caption=false,font=normalsize,labelfont=sf,textfont=sf]{subfig}
\usepackage{textcomp}

\usepackage{url}
\usepackage{verbatim}
\usepackage{graphicx}
\usepackage{cite}
\usepackage{amsmath}
\usepackage{booktabs}
\usepackage{dsfont}
\usepackage{tabularx}
\usepackage{caption}  
\usepackage{dblfloatfix} 
\usepackage{multirow}
\usepackage{orcidlink}

\begin{document}

\title{3DGBGS: 3D Granular Ball Gaussian  Splatting for Compact Novel View Synthesis}

\author{%
    Meng~Yang\orcidlink{0000-0005-8311-092X},~
    Shuyin~Xia\orcidlink{0000-0001-5993-9563},~
    \IEEEmembership{Senior~Member,~IEEE},~
    Dawei~Dai\orcidlink{0000-0002-8431-4431}\textsuperscript{*},~
    and~Yi~Wang%
    \thanks{This work was supported by the Chongqing Key Laboratory of
    Computational Intelligence, the Key Laboratory of Cyberspace Big Data
    Intelligent Security, Ministry of Education, the Sichuan--Chongqing
    Co-construction Key Laboratory of Digital Economy Intelligence, and the
    Key Laboratory of Big Data Intelligent Computing, Chongqing University
    of Posts and Telecommunications, Chongqing, China.}%
    \thanks{Meng Yang, Shuyin Xia, and Dawei Dai are with Chongqing University
    of Posts and Telecommunications, Chongqing, China.}%
    \thanks{Yi Wang is with Chongqing Ant Consumer Finance Co., Ltd.,
    Chongqing, China.}%
    \thanks{\textsuperscript{*}Corresponding author: Dawei Dai
    (dw\_dai@163.com).}%
}

\markboth{Journal of \LaTeX\ Class Files,~Vol.~14, No.~8, August~2021}%
{Shell \MakeLowercase{\textit{et al.}}: A Sample Article Using IEEEtran.cls for IEEE Journals}


\maketitle

\begin{abstract}
Three-dimensional Gaussian Splatting (3DGS) enables high-quality real-time
novel-view synthesis through explicit Gaussian primitives and differentiable
rasterization. 3DGS and Granular Ball Computing, proposed in 2019, share a
natural compatibility in adaptive representation. The efficiency of 3DGS
partly stems from a coarse-to-fine and on-demand refinement process that draws
on the generation principle of Granular Ball Computing. This connection
motivates us to further introduce adaptive granular-ball organization into
anchor-based 3DGS. Existing anchor-based methods typically construct anchors from sparse SfM point
clouds through fixed voxelization, which cannot adequately adapt to spatially
non-uniform point distributions and leads to a trade-off among anchor count,
model compactness, and rendering quality. To address this issue, we propose
\textbf{3DGBGS} (3D Granular Ball Gaussian Splatting), a compact anchor-based
framework for novel-view synthesis. 3DGBGS adaptively partitions SfM point
clouds into 3D granular balls, using larger balls to compactly represent smooth
and redundant regions and smaller balls to preserve complex geometry and local
details. Based on this representation, \textbf{Granular Ball Anchor
Initialization (GBAI)} uses granular ball centers to initialize compact anchor
positions, while the \textbf{Granular Ball Scale Prior (GBSP)} exploits granular
ball radii to provide local scale priors for Gaussian generation. Experiments on four benchmarks show that 3DGBGS reduces initial and final
anchors by 37.1\% and 10.0\%, respectively, and model storage by 9.8\% on
average, while maintaining comparable rendering quality.

\end{abstract}


\section{Introduction}

Three-dimensional scene representation and novel-view synthesis are fundamental
problems in computer vision and computer graphics, with important applications
in virtual reality~\cite{flynn2019deepview,broxton2020immersive}, scene
reconstruction~\cite{mildenhall2020representing,kerbl20233d}, and robotic
perception~\cite{irshad2024neural}. Neural Radiance Fields (NeRFs) model a
scene as an implicit continuous radiance field and achieve high-quality
novel-view synthesis. However, their volumetric rendering process typically
requires extensive sampling and repeated network evaluations, thereby limiting
both training and rendering efficiency~\cite{mildenhall2020representing}. In contrast, 3D Gaussian Splatting (3DGS)~\cite{kerbl20233d} follows the coarse-to-fine and on-demand refinement paradigm of Granular Ball Computing. It initializes coarse Gaussian primitives from sparse SfM points and progressively generates finer primitives through splitting and densification to capture local geometry and appearance details. Combined with explicit Gaussian representation and differentiable rasterization, this adaptive multi-density process enables high-quality real-time novel-view synthesis.

To improve the compactness of 3DGS representations, anchor-based 3DGS methods
introduce anchor structures to organize local Gaussian distributions.
Representative methods such as Scaffold-GS~\cite{lu2024scaffold} typically
construct anchors from Structure-from-Motion (SfM) point clouds
~\cite{snavely2006photo} and predict the opacity, color, scale, rotation, and
other attributes of local neural Gaussian primitives from anchor features and
view-dependent information. Compared with directly optimizing a large number of
independent 3D Gaussians, anchor-based representations enable feature sharing
within local structures, thereby reducing explicit Gaussian redundancy and
achieving a favorable balance between model compactness and rendering quality.
Existing methods commonly construct anchors from SfM point clouds through fixed
voxelization, where the point cloud is quantized into a regular voxel grid and
the centers of non-empty voxels are used as spatial support locations. This
construction is simple and stable, and its effectiveness has been demonstrated
in efficient novel-view synthesis.

Despite these advantages, SfM point clouds of real-world scenes are often
highly non-uniform, and the structural complexity of different local regions
can vary considerably. Consequently, a uniform voxel granularity may not
provide an appropriate spatial organization for all scene regions. Fine-grained
voxelization helps preserve complex geometry, object boundaries, and local
details, but may introduce excessive spatial support units in relatively smooth
regions such as walls, floors, and backgrounds, thereby increasing model size
and storage overhead. Conversely, coarse-grained voxelization can further
reduce the number of anchors, but may weaken the representation of occluded
regions, small-scale structures, and areas with complex textures. Therefore,
constructing flexible multi-granularity spatial supports according to local
point-cloud complexity remains an important problem in compact 3D splatting
representations.

From a more fundamental perspective, compact 3D splatting representations
depend not only on compressing existing Gaussian primitives or anchor features,
but also on how spatially non-uniform SfM point clouds are organized.
Granular-Ball Computing (GBC) was proposed by Xia et al. in
2019~\cite{xia2019granular}. Inspired by the human cognitive principle of
``large-scale first,'' GBC adaptively organizes data into granular balls at
different scales through a coarse-to-fine process. Although granular balls are
typically represented as hyperspherical regions, rectangular regions have been
used for image representation~\cite{shuyin2023graph}, while ellipsoidal
granules have been introduced to better characterize anisotropic data
distributions~\cite{sun2025gec}. By using information granules of different
scales to adaptively cover data and capture their structural relationships,
GBC provides an efficient, robust, and interpretable computational paradigm.
Since its introduction, GBC has been extended to various artificial
intelligence tasks, including granular ball classification
~\cite{xia2021granular,xia2022efficient}, granular ball clustering
~\cite{xia2020ball}, and granular ball rough sets
~\cite{xia2020gbnrs,xia2023gbrs}.

3DGS and Granular Ball computing share many commonalities in adaptive
multi-granularity modeling. One important source of the efficiency of 3DGS lies
in its adoption of the coarse-to-fine and on-demand splitting mechanism of
Granular Ball computing. Granular Ball computing starts from coarse-grained
coverage and recursively partitions granular balls according to local quality
feedback, allowing simple regions to remain at a coarse granularity while
providing finer representations for complex regions
~\cite{Xia2023GranularballCA}. Similarly, 3DGS initializes a
Gaussian set from sparse SfM points and progressively enriches local
representations through cloning or splitting. Moreover, both paradigms use
gradient information to guide local refinement. Granular Ball computing
distinguishes smooth regions from edges and textured regions according to local
gradients, thereby enabling gradient-aware granularity allocation
~\cite{shuyin2023graph}. In contrast, 3DGS uses gradients of the rendering loss
with respect to view-space positions to identify under-represented regions and
selects cloning or splitting according to Gaussian scale. Finally, both
paradigms incorporate redundancy-control mechanisms. Granular Ball methods
suppress ineffective refinement by merging similar granular balls, eliminating
unreasonable overlaps, and introducing termination conditions
~\cite{xia2022gbc,xia2022efficient}, whereas 3DGS prunes Gaussians with low
opacity or abnormal scales. Consequently, both form an efficient adaptive
modeling process characterized by coarse-grained initialization,
gradient-guided refinement, and redundant-unit reduction.

Owing to the natural compatibility between 3DGS and Granular Ball computing,
we introduce granular-ball-based point-cloud organization into anchor-based
3DGS and propose \textbf{3DGBGS}, a compact anchor-based framework for
novel-view synthesis. Specifically, 3DGBGS adaptively partitions a spatially
non-uniform SfM point cloud into 3D granular balls, which serve as region-level
point-cloud organization units. Based on this representation, 
\textbf{Granular Ball Anchor Initialization (GBAI)} uses granular-ball centers
to initialize compact anchor positions, reducing redundant spatial supports
introduced by fixed voxelization in smooth regions. Meanwhile, the
\textbf{Granular Ball Scale Prior (GBSP)} exploits granular-ball radii to
provide local scale priors for subsequent Gaussian generation. By jointly
leveraging GBAI and GBSP, 3DGBGS reduces anchor redundancy while preserving
the spatial supports required for regions with complex geometry and appearance,
thereby improving representation compactness with comparable rendering quality.

The main contributions of this work are summarized as follows:

\begin{itemize}

\item We propose \textbf{3DGBGS}, a compact anchor-based 3D Gaussian
splatting framework that adaptively organizes spatially non-uniform SfM point
clouds into 3D granular balls. Compared with fixed voxelization, this
Granular Ball representation provides structure-aware spatial supports adapted
to local point-cloud distributions.

\item We introduce \textbf{Granular Ball Anchor Initialization (GBAI)} and
the \textbf{Granular Ball Scale Prior (GBSP)}. GBAI uses Granular Ball centers
to initialize compact anchor positions, while GBSP exploits Granular Ball
radii to provide local scale priors, enabling the generated Gaussian primitives
to better conform to local spatial structures.

\item Extensive experiments on 19 scenes from four real-world novel-view
synthesis benchmarks demonstrate that 3DGBGS maintains comparable rendering
quality while reducing the numbers of initial and final anchors by
approximately 37.1\% and 10.0\%, respectively, and model storage by 9.8\%
on average.

\end{itemize}

\section{Related Work}

\subsection{Neural Radiance Fields}

Neural Radiance Fields (NeRF) \cite{mildenhall2020representing} represent a 3D scene as a continuous implicit radiance field learned from multi-view images and camera poses. Given a spatial position and viewing direction, NeRF predicts volumetric density and view-dependent color, and synthesizes novel views through differentiable volume rendering along camera rays. Owing to its strong capability for continuous scene modeling and multi-view consistency, NeRF has been extended to large-scale scenes \cite{barron2022mip}, dynamic reconstruction \cite{pumarola2021d}, sparse-view synthesis \cite{niemeyer2022regnerf}, pose-free reconstruction \cite{lin2021barf}, and anti-aliased rendering \cite{barron2021mip}. However, NeRF-based methods typically require dense ray sampling and repeated network evaluations, resulting in high computational costs during training and rendering. To improve efficiency, subsequent studies have explored explicit feature grids \cite{fridovich2022plenoxels}, multi-resolution hash encoding \cite{muller2022instant}, and tensor decomposition with low-rank representations \cite{chen2022tensorf}. Despite substantial acceleration, these approaches generally retain sampling-based volume rendering, and therefore still face a trade-off among rendering speed, reconstruction quality, and representation compactness.

\subsection{3D Gaussian Splatting}

In recent years, 3D Gaussian Splatting (3DGS)~\cite{kerbl20233d}
has drawn on the efficient coarse-to-fine multi-granularity generation
paradigm of granular-ball computing to generate ellipsoidal primitives of
different sizes. By explicitly representing 3D scenes with Gaussian
primitives, it provides a new technical route for real-time, high-quality
novel-view synthesis. Each Gaussian is parameterized by optimizable
attributes, including position, covariance, opacity, and color, and is
efficiently rendered through differentiable splatting and
$\alpha$-blending. Subsequent studies have extended 3DGS to dynamic scenes
~\cite{luiten2024dynamic,wu20244d}, autonomous driving
~\cite{zhou2024drivinggaussian}, sparse-view reconstruction
~\cite{xiong2023sparsegs}, COLMAP-free reconstruction
~\cite{fu2024colmap}, rendering enhancement
~\cite{yu2024mip}, scene editing
~\cite{wang2024gaussianeditor}, and stylized rendering
~\cite{liu2024stylegaussian}. However, complex scenes often require a large
number of Gaussian primitives, resulting in representation redundancy,
increased storage overhead, and higher optimization costs. Therefore,
developing compact and structure-aware 3D splatting representations remains
an important research direction.

\subsection{Anchor-Based 3D Gaussian Splatting}

Anchor-based 3D Gaussian Splatting organizes local Gaussian primitives around sparse anchors to reduce the redundancy of independently optimized Gaussians. These methods typically construct anchors from an SfM point cloud, associate each anchor with learnable features and local offsets, and use lightweight networks to predict the opacity, color, scale, and rotation of nearby Gaussians. Scaffold-GS \cite{lu2024scaffold} is a representative approach that initializes anchors through fixed voxelization and predicts local Gaussian attributes from anchor features, viewing directions, and viewing distances. SOGS further models second-order correlations among anchor-feature dimensions to reduce feature dimensionality and model size while preserving rendering quality \cite{zhang2025sogs}. Despite these advances, their spatial scaffolds still largely rely on uniform voxel grids, where non-empty voxel centers are used as anchors. Consequently, anchor distribution is mainly determined by a predefined voxel size and cannot adapt well to the non-uniform geometry and point density of real-world SfM point clouds. Unlike existing methods that primarily improve anchor features or Gaussian prediction, this work focuses on spatial scaffold construction and explores a more compact and structure-aware anchor representation based on local point-cloud structures.

\subsection{Granular Ball Computing}

Human cognition generally follows the ``large-range-first'' principle
~\cite{chen1982topological}, whereby coarse-grained global structures are
typically recognized before fine-grained local details. Inspired by this
cognitive mechanism, multi-granularity learning jointly models information at
different scales to capture global layouts and local details in a coordinated
manner~\cite{wang2017dgcc}. In 2019, Xia et al. introduced the fundamental idea of Granular-Ball Computing (GBC)~\cite{xia2019granular}. Subsequently, they further systematized and unified GBC, developing it into an adaptive multi-granularity representation and computation framework~\cite{Xia2023GranularballCA}. Its core idea is to replace the
finest-grained individual samples as the basic inputs and computational units
of a learner with coarse-grained, adaptive multi-granularity information
granules, thereby establishing a unified learning paradigm truly based on
multi-granularity inputs. Since the granularity of granular balls can be
efficiently and adaptively adjusted according to the data distribution, GBC
is capable of fitting arbitrarily complex data distributions in a
multi-granularity manner.

In 2020, GBC was extended to GBNRS~\cite{xia2020gbnrs}. In 2021, GBS and
granular-ball-based feature selection were developed
~\cite{xia2021granular,chen2021granular}. In 2022, subsequent studies
introduced adaptive granular-ball generation~\cite{xia2022efficient},
VPGB~\cite{peng2022vpgb}, granular ball clustering~\cite{xia2022gbc},
GBFS~\cite{xia2022granular}, and GBCD~\cite{chen2022gbcd}. In 2023,
GBC was further extended to GBRS~\cite{xia2023gbrs}, GBPD
~\cite{cheng2023fast}, incremental GBRS~\cite{zhang2023incremental},
GBSC~\cite{xie2023efficient}, and rough granular-ball methods for
multi-label feature selection~\cite{qian2023multi}.

The geometric form of a granular ball is not restricted to an exact
hypersphere. In 2023, rectangular regions were used to approximate granular
balls for image representation while retaining the coarse-to-fine and
quality-controlled partitioning mechanism~\cite{shuyin2023graph}. GBCloud
further combined granular-ball partitioning with Gaussian cloud models
~\cite{shi2023gbcloud}. In November 2023, Kerbl et al. proposed 3DGS~\cite{kerbl20233d}, which also draws on the coarse-to-fine
and representation-quality-driven refinement strategy of GBC. Through
cloning, splitting, and pruning, 3DGS efficiently generates Gaussian
ellipsoidal primitives at multiple scales and employs them for high-quality
real-time rendering.

In 2025, granular-ellipsoid representations were introduced to better
characterize anisotropic data distributions~\cite{sun2025gec}. In 2026,
they were further extended to label-noise filtering, density-peaks
clustering, and adaptive ellipsoid generation
~\cite{zhang2026geaf,liu2026fast,guo2026granular}. Although their geometric
forms differ, these methods still follow the adaptive, coarse-to-fine, and
quality-controlled paradigm of GBC.

Moreover, granular-ball computing has been widely applied to reinforcement
learning~\cite{liu2024unlock}, deep learning
~\cite{shuyin2023graph,10996538,dai2024granular,shen2026finding}, and graph
learning~\cite{xia2025graph,xia2025gbgc}, demonstrating its broad
applicability as a general tool for multi-granularity data representation and
structural modeling.

3DGS and GBC share a natural connection in adaptive multi-granularity
representation. GBC adaptively refines complex regions while retaining coarse
representations in relatively uniform regions, whereas 3DGS generates Gaussian
primitives at different scales through densification, splitting, and pruning.
Motivated by this connection, we organize spatially non-uniform SfM point
clouds using adaptive 3D granular balls. Their centers initialize compact
anchor positions, while their radii provide local scale priors. Larger
granular balls compactly cover smooth and redundant regions, whereas smaller
ones preserve geometric and appearance details in structurally complex
regions, resulting in a compact and structure-aware point-cloud
representation.

\section{Preliminaries}
\label{sec:preliminaries}

\subsection{Granular Ball Computing}
\label{sec:pre_gbc}

Granular Ball Computing (GBC)~\cite{Xia2023GranularballCA} is an adaptive multi-granularity representation and learning paradigm. Unlike conventional sample-wise learning methods that treat individual samples as the basic processing units, GBC aggregates samples with similar distributions or consistent characteristics into granularity units and represents the original data space using a set of granular balls. Its central objective is to balance coverage, compactness, the number of granular balls, and granular ball quality. On the one hand, the granular ball set should cover the original sample space as completely as possible to avoid information loss. On the other hand, samples within each granular ball should maintain sufficient consistency so that the ball retains meaningful local semantic or structural information.

Given a sample set
\begin{equation}
\mathcal{D}
=
\left\{
\mathbf{z}_{1},
\mathbf{z}_{2},
\ldots,
\mathbf{z}_{n}
\right\},
\end{equation}
GBC aims to construct a granular ball set
\begin{equation}
\mathcal{G}
=
\left\{
GB_{1},
GB_{2},
\ldots,
GB_{m}
\right\},
\end{equation}
to cover and represent the original dataset $\mathcal{D}$. The $i$-th granular ball is generally represented as
\begin{equation}
GB_i
=
\left(
\mathcal{D}_i,
\mathbf{c}_i,
r_i,
q_i
\right),
\end{equation}
where $\mathcal{D}_i\subseteq\mathcal{D}$ denotes the subset of samples contained in the $i$-th granular ball, $\mathbf{c}_i$ and $r_i$ denote its center and radius, respectively, and $q_i$ denotes a quality measure, such as class purity, geometric compactness, or feature consistency.

In general, the sample subsets associated with different granular balls form a partition of the original sample set:
\begin{equation}
\mathcal{D}
=
\bigcup_{i=1}^{m}
\mathcal{D}_i,
\qquad
\mathcal{D}_i
\cap
\mathcal{D}_j
=
\emptyset,
\quad
i\neq j .
\end{equation}

From an optimization perspective, granular ball construction can be interpreted as balancing coverage, compactness, granular ball quantity, and granular ball quality. A representative granular ball formulation can be written as
\begin{equation}
\begin{aligned}
\min_{\mathcal{G}}\quad&
-\operatorname{Cov}(\mathcal{G})
-\operatorname{Comp}(\mathcal{G})
+m,\\
\mathrm{s.t.}\quad&
\operatorname{quality}(GB_i)\geq\Phi(\mathcal{D}),
\quad i=1,\ldots,m.
\end{aligned}
\end{equation}
Here, $\operatorname{Cov}(\mathcal{G})$ measures the coverage of the original
sample space, $\operatorname{Comp}(\mathcal{G})$ measures intra-ball
compactness, $m$ denotes the number of granular balls,
$\operatorname{quality}(GB_i)$ denotes the quality of the $i$-th granular
ball, and $\Phi(\mathcal{D})$ specifies the corresponding quality constraint
or threshold. The objective is to represent the original data with as few
granular balls as possible while maintaining sufficient coverage and
intra-ball consistency. In practice, coverage, compactness, and ball quality
are controlled by predefined parameters, thresholds, or constraints, whereas
$m$ is minimized through a coarse-to-fine, coarse-granularity-first strategy:
the data space is first covered by coarse granular balls, and only those that
fail to satisfy the quality requirement are further split, thereby avoiding
unnecessary fine-grained balls.





\subsection{3D Gaussian Splatting}
\label{sec:pre_3dgs}

3D Gaussian Splatting (3DGS)~\cite{kerbl20233d} explicitly represents a
three-dimensional scene as a set of anisotropic 3D Gaussians. The $i$-th
Gaussian is typically parameterized by its center position $\mathbf{x}_i$,
opacity $\alpha_i$, color $\mathbf{c}_i$, scale vector $\mathbf{s}_i$, and
rotation quaternion $\mathbf{q}_i$. The scale and rotation jointly determine
its covariance matrix:
\begin{equation}
\boldsymbol{\Sigma}_i
=
\mathbf{R}(\mathbf{q}_i)
\operatorname{diag}
\left(
\mathbf{s}_i^{2}
\right)
\mathbf{R}(\mathbf{q}_i)^{\top},
\end{equation}
where $\mathbf{R}(\mathbf{q}_i)$ denotes the rotation matrix derived from the
rotation quaternion. The covariance matrix $\boldsymbol{\Sigma}_i$ jointly
characterizes the spatial scale, anisotropic shape, and orientation of the
Gaussian primitive.

From the representation perspective of granular-ball computing, a 3D Gaussian
ellipsoid can be modeled as a generalized granular-ball unit. Specifically,
the $i$-th Gaussian ellipsoid can be represented as
\begin{equation}
\widetilde{GB}_i
=
\left(
\Omega_i,
\mathbf{x}_i,
\boldsymbol{\Sigma}_i,
\rho_i
\right),
\end{equation}
where $\Omega_i$ denotes the local spatial region covered by the Gaussian
primitive and can be defined by its covariance as
\begin{equation}
\Omega_i
=
\left\{
\mathbf{p}
\ \middle|\
(\mathbf{p}-\mathbf{x}_i)^{\top}
\boldsymbol{\Sigma}_i^{-1}
(\mathbf{p}-\mathbf{x}_i)
\leq \tau
\right\},
\end{equation}
$\mathbf{x}_i$ denotes the center of the spatial unit,
$\boldsymbol{\Sigma}_i$ describes its anisotropic spatial extent, and
$\rho_i$ denotes the local importance or representation quality characterized
by information such as opacity, rendering contribution, or gradient feedback.
Compared with the scalar radius $r_i$ used in conventional granular-ball
models, $\boldsymbol{\Sigma}_i$ generalizes the scale representation to an
orientation-aware anisotropic form. Therefore, a 3DGS scene can be uniformly
represented as a set of generalized granular balls with different positions,
scales, and orientations:
\begin{equation}
\widetilde{\mathcal{G}}
=
\left\{
\widetilde{GB}_i
\right\}_{i=1}^{M}.
\end{equation}

During rendering, the 3D Gaussians are first projected onto the 2D image plane,
after which pixel colors are computed through splatting and
$\alpha$-blending. For a given pixel, its color $\mathbf{C}$ can be written as
\begin{equation}
\mathbf{C}
=
\sum_{n\in\mathcal{N}}
\mathbf{c}_{n}
\alpha_{n}
\prod_{z=1}^{n-1}
(1-\alpha_{z}),
\end{equation}
where $\mathcal{N}$ denotes the set of projected 2D Gaussians that cover the
pixel and are sorted by depth, while $\mathbf{c}_{n}$ and $\alpha_{n}$ denote
the color and opacity of the $n$-th Gaussian, respectively.

3DGS first generates and progressively refines Gaussian ellipsoidal primitives
in an efficient coarse-to-fine manner. Based on these explicit Gaussian
representations, it then achieves high-quality, real-time novel-view synthesis
through differentiable rasterization. However, since the original 3DGS
directly maintains a large number of independent 3D Gaussians, redundant
representations can easily arise in large-scale scenes or smoothly textured
regions, leading to substantial storage and optimization overhead.

\begin{figure*}[t]
  \centering
  \includegraphics[width=1\linewidth]{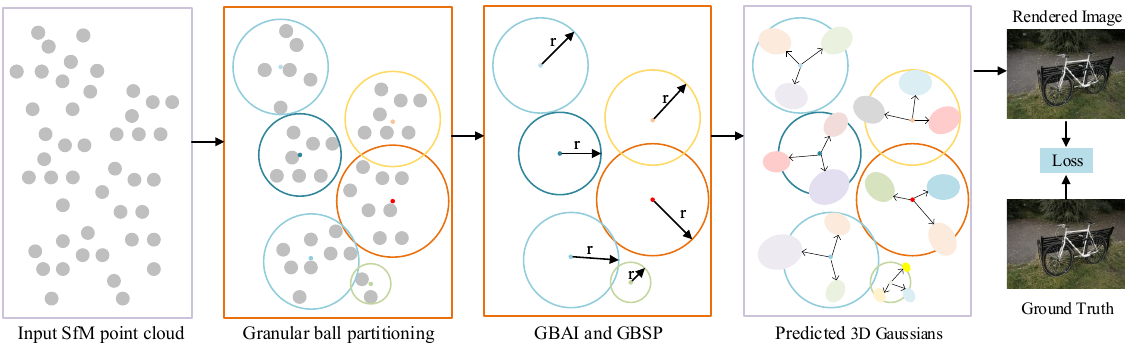}
  \caption{
  \textbf{Overall framework of 3DGBGS.}
  The spatially non-uniform SfM point cloud is first adaptively partitioned
  into 3D granular balls. Granular Ball Anchor Initialization (GBAI) uses
  their centers to initialize compact anchor positions, while the
  Granular Ball Scale Prior (GBSP) exploits their radii to provide local
  scale priors. Local Gaussian primitives are then generated from the
  Granular Ball-guided anchors and rendered through differentiable splatting.
  The model is optimized using reconstruction losses between the rendered
  and ground-truth images.
  }
  \label{fig:framework}
\end{figure*}

\subsection{Scaffold-GS}
\label{sec:pre_scaffold}

Scaffold-GS~\cite{lu2024scaffold} improves 3DGS by introducing an anchor scaffold and a neural Gaussian prediction mechanism. Instead of directly optimizing a large number of independent 3D Gaussians, Scaffold-GS first constructs a set of anchors from an SfM point cloud and then predicts local neural Gaussians around each anchor, thereby reducing Gaussian redundancy and improving representation efficiency.

Given an SfM point cloud
\begin{equation}
\mathcal{P}=\{\mathbf{x}_{i}\}_{i=1}^{N}
\end{equation}
and a voxel size $v$, Scaffold-GS constructs the anchor set through fixed voxelization:
\begin{equation}
\mathcal{A}_{\mathrm{vox}}
=
\left\{
v\cdot
\mathrm{round}
\left(
\frac{\mathbf{x}_{i}}{v}
\right)
\mid
\mathbf{x}_{i}\in\mathcal{P}
\right\}_{\mathrm{unique}},
\end{equation}
where $\{\cdot\}_{\mathrm{unique}}$ denotes the removal of duplicated voxel centers.

For each anchor $\mathbf{x}_{a}\in\mathcal{A}_{\mathrm{vox}}$, Scaffold-GS maintains a learnable anchor feature $\mathbf{f}^{a}\in\mathbb{R}^{D}$, an anchor scale $\mathbf{l}^{a}$, and $K$ learnable offsets $\{\mathbf{o}^{a}_{k}\}_{k=1}^{K}$. Given a camera center $\mathbf{x}_{c}$, the distance and viewing direction from the anchor to the camera are defined as
\begin{equation}
d_{ac}
=
\left\|
\mathbf{x}_{a}
-
\mathbf{x}_{c}
\right\|_{2},
\qquad
\boldsymbol{\delta}_{ac}
=
\frac{
\mathbf{x}_{a}
-
\mathbf{x}_{c}
}{
\left\|
\mathbf{x}_{a}
-
\mathbf{x}_{c}
\right\|_{2}
}.
\end{equation}

Lightweight MLPs then predict the attributes of $K$ local Gaussians from the anchor feature, viewing direction, and viewing distance:
\begin{equation}
\left\{
\alpha_{k},
\mathbf{c}_{k},
\mathbf{s}_{k},
\mathbf{q}_{k}
\right\}_{k=1}^{K}
=
F
\left(
\mathbf{f}^{a},
\boldsymbol{\delta}_{ac},
d_{ac}
\right),
\end{equation}
where $\alpha_{k}$, $\mathbf{c}_{k}$, $\mathbf{s}_{k}$, and $\mathbf{q}_{k}$ denote the opacity, color, scale, and rotation parameters of the $k$-th Gaussian, respectively.

The position of the $k$-th Gaussian generated from anchor $\mathbf{x}_{a}$ is given by
\begin{equation}
\mathbf{p}_{k}
=
\mathbf{x}_{a}
+
\mathbf{o}^{a}_{k}
\odot
\mathbf{l}^{a}_{xyz},
\end{equation}
where $\odot$ denotes element-wise multiplication and $\mathbf{l}^{a}_{xyz}$ denotes the first three spatial dimensions of the anchor scale.

The regularized anchor construction of Scaffold-GS is simple and effective, providing stable spatial supports for efficient neural Gaussian prediction. Building upon this framework, we further investigate the organization of the point cloud before Gaussian generation. Specifically, we employ granular balls to model the non-uniform SfM point cloud at multiple granularities, thereby providing more compact and structure-aware spatial supports together with local scale priors.

\section{Proposed Method}

We propose \textbf{3DGBGS}, a compact 3D Granular Ball Gaussian splatting
framework for novel-view synthesis. Instead of constructing anchors with fixed
voxel grids, 3DGBGS adaptively partitions non-uniform SfM point clouds into
multi-granularity 3D granular balls. Large balls compactly represent smooth and
redundant regions, while small balls preserve complex geometry and local
details. Their centers initialize anchor positions, and their radii provide
local scale priors, thereby reducing redundant anchors while retaining the
spatial supports required in complex regions. The overall framework is shown in
Fig.~\ref{fig:framework}.

\subsection{Adaptive Granular Ball Construction and Anchor Initialization}
\label{sec:gb_construction}

\paragraph{Granular Ball Construction}
Anchor-based 3DGS methods generally require spatial support units to be constructed from an SfM point cloud before Gaussian generation. Existing methods commonly quantize the point cloud into a regular voxel grid through fixed voxelization and use the centers of non-empty voxels as anchors. Although simple and stable, this fixed-granularity strategy cannot adequately accommodate the spatial non-uniformity of real-world SfM point clouds. Smooth or redundant regions may produce excessive and highly similar anchors, whereas complex geometry, occlusion boundaries, and regions with significant appearance variation require finer local supports.

To address this limitation, 3DGBGS introduces granular ball-based point-cloud organization before Gaussian generation and partitions the non-uniform SfM point cloud into a set of 3D granular balls. Unlike fixed voxel centers, granular balls adapt their spatial scales according to local geometric distributions and appearance consistency. Larger granular balls compactly cover structurally consistent regions, whereas regions exhibiting substantial structural variation are recursively partitioned into finer-grained granular balls to preserve local details. The resulting granular ball set therefore provides a compact and structure-aware spatial basis for subsequent Gaussian generation.

Given an SfM point cloud
\begin{equation}
\mathcal{P}
=
\left\{
(\mathbf{x}_{i},\mathbf{a}_{i})
\right\}_{i=1}^{N},
\end{equation}
where $\mathbf{x}_{i}\in\mathbb{R}^{3}$ denotes the 3D coordinate of the $i$-th point and $\mathbf{a}_{i}\in\mathbb{R}^{3}$ denotes its appearance attribute, 3DGBGS partitions the point cloud into $M$ mutually disjoint local point sets:
\begin{equation}
\mathcal{P}
=
\bigcup_{j=1}^{M}
\mathcal{P}_{j},
\qquad
\mathcal{P}_{j}
\cap
\mathcal{P}_{k}
=
\emptyset,
\quad
j\neq k .
\end{equation}
Each local point set $\mathcal{P}_{j}$ corresponds to a 3D granular ball:
\begin{equation}
GB_j
=
\left(
\mathcal{P}_{j},
\boldsymbol{\mu}_{j},
r_{j}
\right),
\end{equation}
where $\boldsymbol{\mu}_{j}$ and $r_j$ denote the center and radius of the granular ball, respectively.

The center of the $j$-th granular ball is defined as the mean coordinate of its constituent points:
\begin{equation}
\boldsymbol{\mu}_{j}
=
\frac{1}
{|\mathcal{P}_{j}|}
\sum_{(\mathbf{x}_{i},\mathbf{a}_{i})\in\mathcal{P}_{j}}
\mathbf{x}_{i}.
\end{equation}

To characterize the spatial support range of the local point set, we define the granular ball radius using the root-mean-square distance:
\begin{equation}
r_{j}
=
\max
\left(
\sqrt{
\frac{1}
{|\mathcal{P}_{j}|}
\sum_{(\mathbf{x}_{i},\mathbf{a}_{i})\in\mathcal{P}_{j}}
\left\|
\mathbf{x}_{i}
-
\boldsymbol{\mu}_{j}
\right\|_{2}^{2}
},
s_{\min}
\right).
\end{equation}
Here, $s_{\min}$ is a minimum-scale constraint that prevents singleton or extremely small granular balls from producing degenerate radii. In our implementation, it is defined as
\begin{equation}
s_{\min}
=
\max(\gamma v,\epsilon_s),
\end{equation}
where $v$ denotes the base spatial scale, $\gamma$ is a minimum-scale ratio, and $\epsilon_s$ is a small positive constant.

To adapt the granular balls to the local structural distributions of the SfM point cloud, we construct the granular ball set using recursive binary partitioning. For the current point set $\mathcal{P}_{j}$, we first calculate its spatial radius $r_j$ and appearance variation:
\begin{equation}
\sigma_{a}(\mathcal{P}_{j})
=
\frac{1}{3}
\sum_{m=1}^{3}
\sqrt{
\frac{1}
{|\mathcal{P}_{j}|}
\sum_{(\mathbf{x}_{i},\mathbf{a}_{i})\in\mathcal{P}_{j}}
\left(
a_{i}^{m}
-
\bar{a}_{j}^{m}
\right)^{2}
},
\end{equation}
where $a_i^m$ denotes the value of the $i$-th point in the $m$-th appearance channel, and $\bar{a}_{j}^{m}$ denotes the mean value of $\mathcal{P}_{j}$ in that channel.

The partitioning process terminates when any of the following conditions is satisfied:
\begin{equation}
|\mathcal{P}_{j}|
\leq
N_{\min}
\quad
\lor
\quad
d_j
\geq
D_{\max}
\quad
\lor
\quad
\left(
r_j
\leq
\eta v
\ \land\
\sigma_{a}(\mathcal{P}_{j})
\leq
\tau_a
\right),
\end{equation}
where $N_{\min}$ denotes the minimum number of points permitted in a granular ball, $d_j$ denotes the current recursion depth, $D_{\max}$ is the maximum recursion depth, $\eta$ is a radius-control coefficient, and $\tau_a$ is an appearance-consistency threshold. This stopping criterion treats the current point set as a valid granular ball when it is sufficiently small, the maximum recursion depth has been reached, or the point set is both spatially compact and appearance-consistent. Otherwise, the point set is further divided into smaller subsets.

Specifically, we adopt a PCA-based binary partitioning strategy. We first compute the coordinate covariance matrix of the current point set:
\begin{equation}
\mathbf{C}_{j}
=
\frac{1}
{|\mathcal{P}_{j}|-1}
\sum_{(\mathbf{x}_{i},\mathbf{a}_{i})\in\mathcal{P}_{j}}
\left(
\mathbf{x}_{i}
-
\boldsymbol{\mu}_{j}
\right)
\left(
\mathbf{x}_{i}
-
\boldsymbol{\mu}_{j}
\right)^{\top}.
\end{equation}
The eigenvector associated with the largest eigenvalue is selected as the principal partitioning direction $\mathbf{u}_{j}$, and each point is projected onto this direction:
\begin{equation}
z_i
=
\left(
\mathbf{x}_{i}
-
\boldsymbol{\mu}_{j}
\right)^{\top}
\mathbf{u}_{j}.
\end{equation}

The current point set is then divided into two subsets according to the median projected value:
\begin{equation}
\mathcal{P}_{j}^{-}
=
\left\{
(\mathbf{x}_{i},\mathbf{a}_{i})\in\mathcal{P}_{j}
\mid
z_i
\leq
\operatorname{median}(\{z_i\})
\right\},
\end{equation}
\begin{equation}
\mathcal{P}_{j}^{+}
=
\left\{
(\mathbf{x}_{i},\mathbf{a}_{i})\in\mathcal{P}_{j}
\mid
z_i
>
\operatorname{median}(\{z_i\})
\right\}.
\end{equation}
If the partition produces an empty subset or a subset violating the minimum-point constraint, the current point set is retained as a terminal granular ball. This PCA-based strategy recursively partitions the local point cloud along its principal spatial-distribution direction, resulting in a structure-adaptive set of multi-granularity granular balls.

\paragraph{Granular Ball Anchor Initialization (GBAI)}
\label{sec:gb_anchor}

After constructing the Granular Ball set, 3DGBGS employs
\textbf{Granular Ball Anchor Initialization (GBAI)} to convert each
Granular Ball center into an anchor position:
\begin{equation}
\mathcal{A}_{\mathrm{GB}}
=
\left\{
\mathbf{x}^{gb}_{j}
=
\boldsymbol{\mu}_{j}
\right\}_{j=1}^{M}.
\end{equation}

Unlike the fixed voxel anchor set $\mathcal{A}_{\mathrm{vox}}$, the
Granular Ball anchor set $\mathcal{A}_{\mathrm{GB}}$ is adaptively determined
by the local distribution of the SfM point cloud. In smooth or redundant
regions, larger granular balls cover more points with fewer anchors. In
regions with complex geometry or significant appearance variations, smaller
granular balls provide finer spatial supports to preserve local details.
Consequently, GBAI produces a compact and structure-aware anchor layout that
better adapts to the spatially non-uniform SfM point cloud.

\begin{figure*}[!htbp]
  \centering
  \includegraphics[width=1\linewidth]{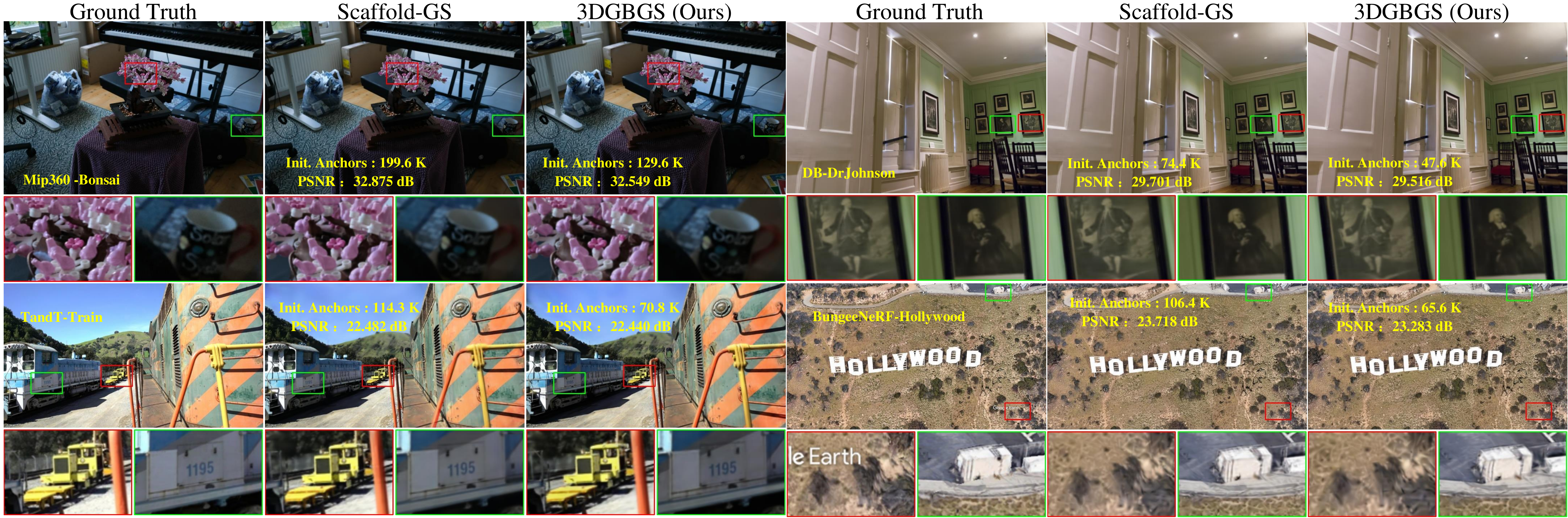}
  \caption{
  Qualitative comparisons of 3DGBGS (ours) with Scaffold-GS~\cite{lu2024scaffold}. 
  }
  \label{fig:visual_comparison}
\end{figure*}

\subsection{Granular Ball Scale Prior}
\label{sec:gb_scale}

In addition to anchor positions, the initial scale determines the spatial
extent of the Gaussian primitives generated from each anchor. We therefore
introduce the \textbf{Granular Ball Scale Prior (GBSP)}, which incorporates
Granular Ball radii into scale initialization.

For the $j$-th Granular Ball anchor $\mathbf{x}^{gb}_{j}$, we first compute a
base scale from its nearest-neighbor distance:
\begin{equation}
d^{\mathrm{nn}}_{j}
=
\sqrt{
\min_{k\neq j}
\left\|
\mathbf{x}^{gb}_{j}
-
\mathbf{x}^{gb}_{k}
\right\|_{2}^{2}
}.
\end{equation}

The scalar distance is then expanded into a six-dimensional scale vector:
\begin{equation}
\mathbf{s}^{\mathrm{nn}}_{j}
=
d^{\mathrm{nn}}_{j}
\mathbf{1}_{6},
\end{equation}
where $\mathbf{1}_{6}$ denotes a six-dimensional all-ones vector.

Although the nearest-neighbor distance characterizes the spacing between
anchors, it does not directly describe the local point-cloud extent represented
by each anchor. In contrast, the Granular Ball radius $r_j$ naturally reflects
the spatial coverage of the corresponding local region. We therefore define
the Granular Ball scale prior as
\begin{equation}
\mathbf{s}^{\mathrm{gb}}_{j}
=
r_j\mathbf{1}_{6}.
\end{equation}

Because Granular Ball radii may vary substantially across different regions,
directly using $\mathbf{s}^{\mathrm{gb}}_{j}$ may produce excessively small or
large initial scales. To stabilize initialization, we constrain the
radius-based prior using the nearest-neighbor scale:
\begin{equation}
\hat{\mathbf{s}}^{\mathrm{gb}}_{j}
=
\operatorname{clip}
\left(
\mathbf{s}^{\mathrm{gb}}_{j},
\lambda_{\min}\mathbf{s}^{\mathrm{nn}}_{j},
\lambda_{\max}\mathbf{s}^{\mathrm{nn}}_{j}
\right),
\end{equation}
where $\lambda_{\min}$ and $\lambda_{\max}$ define the lower and upper bounds,
respectively.

The final initial scale is obtained by fusing the nearest-neighbor scale with
the constrained Granular Ball prior:
\begin{equation}
\mathbf{s}^{0}_{j}
=
\alpha\mathbf{s}^{\mathrm{nn}}_{j}
+
(1-\alpha)\hat{\mathbf{s}}^{\mathrm{gb}}_{j},
\end{equation}
where $\alpha$ balances inter-anchor spacing and local point-cloud coverage.

The fused scale is finally converted into a learnable log-scale parameter:
\begin{equation}
\mathbf{l}^{gb}_{j}
=
\log
\left(
\max
\left(
\mathbf{s}^{0}_{j},
\epsilon
\right)
\right),
\end{equation}
where $\epsilon$ is a small positive constant for numerical stability. The
resulting parameter initializes the scale of subsequent Gaussian primitives
and remains learnable during training.

By combining nearest-neighbor spacing with Granular Ball coverage, GBSP aligns
the initial Gaussian scales with local point-cloud structures more effectively
than relying solely on inter-anchor distances.

\begin{table*}[!htbp]
\centering
\caption{
Quantitative comparison with Scaffold-GS on four real-world novel-view
synthesis benchmarks. Both methods use the same input data, training settings,
rendering pipeline, and dynamic anchor-adjustment strategy. Results are
averaged over all scenes in each dataset. HAC~\cite{chen2024hac} and
ContextGS~\cite{wang2024contextgs} are excluded because they primarily reduce
serialized storage through compression, whereas this comparison focuses on
the active anchors used during training and rendering.
}
\label{tab:main_results}
\setlength{\tabcolsep}{4.5pt}
\renewcommand{\arraystretch}{1.12}
\resizebox{\textwidth}{!}{
\begin{tabular}{l|l|ccc|c|ccc}
\toprule
Dataset & Method 
& SfM Points (K)
& Init. Anchors (K)$\downarrow$
& Final Anchors (K)$\downarrow$
& Storage (MB)$\downarrow$
& PSNR$\uparrow$
& SSIM$\uparrow$
& LPIPS$\downarrow$ \\
\midrule
\multirow{2}{*}{Mip-NeRF360}
& Scaffold-GS 
& 114.7 & 109.8 & 567.2 & 171.6 
& \textbf{27.761} & \textbf{0.812} & \textbf{0.226} \\
& 3DGBGS (Ours)
& 114.7 & \textbf{66.1} & \textbf{521.8} & \textbf{158.0} 
& 27.597 & 0.809 & 0.230 \\
\midrule
\multirow{2}{*}{Tanks\&Temples}
& Scaffold-GS 
& 159.4 & 132.4 & 255.8 & 77.5 
& \textbf{24.136} & \textbf{0.852} & \textbf{0.175} \\
& 3DGBGS (Ours)
& 159.4 & \textbf{92.7} & \textbf{228.1} & \textbf{69.5} 
& 24.081 & 0.850 & 0.180 \\
\midrule
\multirow{2}{*}{Deep Blending}
& Scaffold-GS 
& 58.9 & 54.8 & 175.9 & 53.5 
& \textbf{30.242} & \textbf{0.906} & \textbf{0.256} \\
& 3DGBGS (Ours)
& 58.9 & \textbf{34.1} & \textbf{160.7} & \textbf{49.0} 
& 30.123 & 0.905 & 0.257 \\
\midrule
\multirow{2}{*}{BungeeNeRF}
& Scaffold-GS 
& 117.4 & 113.5 & 579.7 & 175.2 
& 24.891 & 0.837 & \textbf{0.202} \\
& Ours
& 117.4 & \textbf{65.2} & \textbf{510.2} & \textbf{154.3} 
& \textbf{24.946} & \textbf{0.842} & 0.211 \\
\bottomrule
\end{tabular}
}
\end{table*}

\subsection{Optimization and Rendering}
\label{sec:optimization}

After Granular Ball Anchor Initialization (GBAI) and the Granular Ball
Scale Prior (GBSP), 3DGBGS follows the Gaussian prediction, differentiable
rendering, and optimization pipeline of Scaffold-GS~\cite{lu2024scaffold}.
Specifically, each granular ball-guided anchor maintains a learnable feature,
scale parameter, and a set of local offsets. Given the camera viewpoint,
the anchor features and view-dependent information are fed into the original
MLPs to predict the opacity, color, scale, and rotation of the associated
local Gaussian primitives. These Gaussians are subsequently rendered through
differentiable splatting.

The training objective is identical to that of Scaffold-GS:
\begin{equation}
\mathcal{L}
=
\lambda_{1}\mathcal{L}_{1}
+
\lambda_{\mathrm{DSSIM}}\mathcal{L}_{\mathrm{DSSIM}}
+
\lambda_{\mathrm{vol}}\mathcal{L}_{\mathrm{vol}},
\end{equation}
where $\mathcal{L}_{1}$ and $\mathcal{L}_{\mathrm{DSSIM}}$ denote the
pixel-wise reconstruction and structural similarity losses, respectively,
and $\mathcal{L}_{\mathrm{vol}}$ regularizes the spatial extent of the Gaussian
primitives. All loss weights are kept identical to those of Scaffold-GS.

We also retain the original dynamic anchor growing and pruning strategy.
Anchors are added to regions that require additional representation capacity,
while anchors with limited contributions are removed. Therefore, the proposed
method modifies only the initial anchor organization and scale prior, while
leaving the subsequent optimization and rendering procedures unchanged.

\section{Experiments}
\label{sec:experiments}

We evaluate 3DGBGS on multiple real-world novel-view synthesis benchmarks to investigate whether multi-granularity point-cloud organization can reduce the number of spatial support units and model storage while preserving rendering quality. We report rendering quality, initial and final anchor counts, and model storage, followed by hyperparameter and component-wise ablation studies.

\subsection{Experimental Setup}
\label{sec:experimental_setup}

\paragraph{Datasets}
We evaluate 3DGBGS on 19 scenes from four real-world novel-view synthesis
benchmarks. Mip-NeRF 360~\cite{barron2022mip} contains nine indoor and
outdoor scenes with complex camera trajectories and varying scene scales.
Tanks\&Temples~\cite{knapitsch2017tanks} provides the \textit{Truck} and
\textit{Train} scenes with detailed object geometry. Deep Blending
~\cite{hedman2018deep} contains two indoor scenes, \textit{DrJohnson} and
\textit{Playroom}, while BungeeNeRF~\cite{xiangli2022bungeenerf} includes six
large-scale outdoor scenes with substantial scale variations. We follow
Scaffold-GS in data splits, image resolutions, and evaluation protocols.

\paragraph{Evaluation Metrics}
Rendering quality is evaluated using PSNR~\cite{huynh2008scope},
SSIM~\cite{wang2004image}, and LPIPS~\cite{zhang2018unreasonable}, where higher
PSNR and SSIM and lower LPIPS indicate better quality. To assess representation
compactness, we report the numbers of input SfM points, initial anchors, and
final anchors after dynamic growing and pruning, together with model storage.
Storage is measured as the total size of the learned anchor parameters and MLP
weights.

\begin{table*}[!htbp]
\centering
\caption{
Average ablation results on BungeeNeRF~\cite{xiangli2022bungeenerf}.
Point and anchor counts are reported in thousands, and storage is reported in MB.
}
\label{tab:ablation_bungee_avg}
\setlength{\tabcolsep}{4.5pt}
\renewcommand{\arraystretch}{1.12}
\resizebox{\linewidth}{!}{
\begin{tabular}{l|cccc|ccc}
\toprule
Method 
& SfM Points (K)
& Init. Anchors (K)$\downarrow$
& Final Anchors (K)$\downarrow$
& Storage (MB)$\downarrow$
& PSNR$\uparrow$
& SSIM$\uparrow$
& LPIPS$\downarrow$ \\
\midrule
Scaffold-GS 
& 117.4K & 113.5K & 579.7K & 175.2
& 24.891 & 0.837 & 0.202 \\
+ GBAI
& 117.4K & 65.2K & 547.4K & 165.5
& 24.962 & 0.834 & 0.216 \\
+ GBAI + GBSP
& 117.4K & 65.2K & 510.2K & 154.3
& 24.946 & 0.842 & 0.211 \\
\bottomrule
\end{tabular}
}
\end{table*}

\paragraph{Implementation Details}
3DGBGS is implemented on Scaffold-GS~\cite{lu2024scaffold}. The MLP
architecture, feature dimensions, number of offsets, optimizer, training
iterations, reconstruction losses, rendering pipeline, and dynamic anchor
adjustment strategy are kept unchanged. The modifications are limited to
Granular Ball Anchor Initialization (GBAI) and the Granular Ball Scale Prior
(GBSP).

For Granular Ball construction, we set
$N_{\min}=2$, $D_{\max}=18$, $\eta=1.25$, and $\tau_a=0.02$.
The base spatial scale $v$ follows the voxel size of the baseline; when it is
not specified, $v$ is estimated using the median nearest-neighbor distance of
the SfM point cloud. The minimum Granular Ball radius is set to
$s_{\min}=\max(0.25v,10^{-6})$.

For GBSP, the radius-based scale prior is constrained within
$[0.25\mathbf{s}^{\mathrm{nn}},\,4.0\mathbf{s}^{\mathrm{nn}}]$ and fused with
the nearest-neighbor scale using $\alpha=0.5$. Unless otherwise specified, all
remaining training, rendering, growing, pruning, and evaluation settings are
identical to those of Scaffold-GS.

\subsection{Comparison with Anchor-Based 3DGS}
\label{sec:sota}

Because 3DGBGS modifies anchor construction and scale initialization before
Gaussian generation, Scaffold-GS~\cite{lu2024scaffold} is adopted as the
direct baseline. Both methods use identical input SfM point clouds, training
settings, rendering pipelines, and dynamic anchor growing and pruning
strategies. Their differences are limited to Granular Ball Anchor
Initialization (GBAI) and the Granular Ball Scale Prior (GBSP).

As shown in Table~\ref{tab:main_results}, 3DGBGS reduces the number of initial
anchors by 30.0\%--42.6\% across the four benchmarks, demonstrating that
adaptive granular ball partitioning provides a substantially more compact
initialization than fixed voxelization. This advantage is largely preserved
after optimization, with reductions of 8.0\%--12.0\% in final anchors and
7.9\%--11.9\% in model storage. Meanwhile, the differences in rendering
quality remain limited. In particular, on BungeeNeRF, 3DGBGS improves PSNR
and SSIM while reducing the final anchor count and storage by 12.0\% and
11.9\%, respectively.

The qualitative results in Fig.~\ref{fig:visual_comparison} further show that
3DGBGS preserves fine structures, object boundaries, and local appearance
despite using substantially fewer initial anchors. Overall, 3DGBGS reduces
the initial and final anchor counts by 37.1\% and 10.0\%, respectively, and
model storage by 9.8\% on average, while maintaining rendering quality
comparable to Scaffold-GS.

\subsection{Ablation Study}
\label{sec:ablation}

Table~\ref{tab:ablation_bungee_avg} reports the average ablation results on
BungeeNeRF. Replacing voxel-based initialization with GBAI reduces the number
of initial anchors from 113.5K to 65.2K, corresponding to a 42.5\% reduction,
while also decreasing the final anchor count and model storage. Adding GBSP
does not change the initial anchor count, but further reduces the final anchor
count from 547.4K to 510.2K and storage from 165.5 MB to 154.3 MB. Compared
with Scaffold-GS, the complete model reduces the initial and final anchor
counts by 42.5\% and 12.0\%, respectively, and model storage by 11.9\%.
Meanwhile, it achieves slightly higher PSNR and SSIM, with only a minor LPIPS
trade-off. These results show that GBAI mainly provides compact anchor
initialization, while GBSP further suppresses unnecessary anchor growth during
optimization.

\begin{figure*}[!htbp]
  \centering
  \includegraphics[width=1\linewidth]{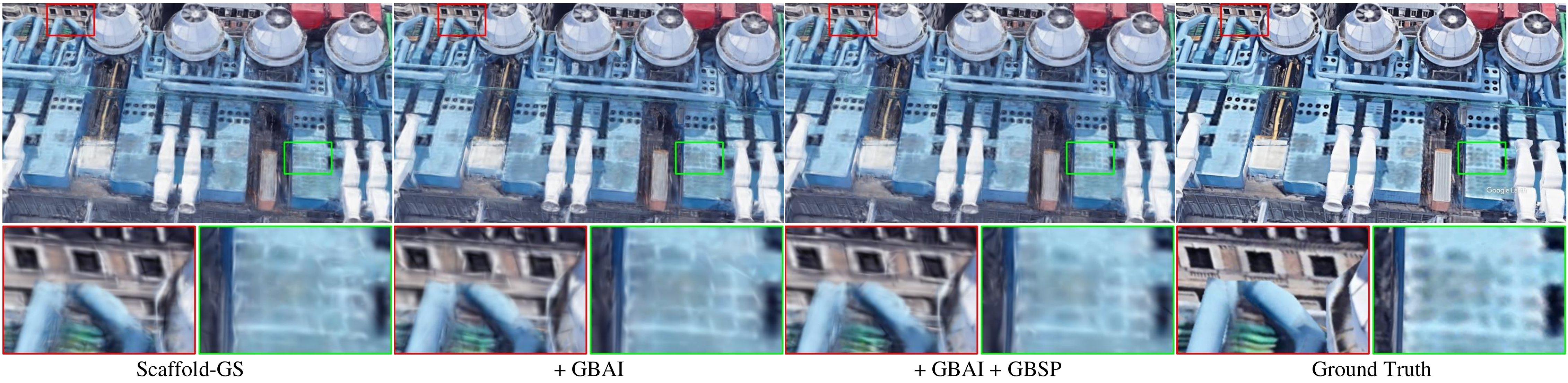}
  \caption{
    Visual ablation comparison of Scaffold-GS, GBAI, and the complete model.
    The highlighted regions show that GBAI preserves the overall scene structure,
    while GBSP further improves local detail consistency under a more compact
    anchor representation.
  }
  \label{fig:ablation_visual}
\end{figure*}

As shown in Fig.~\ref{fig:ablation_visual}, GBAI preserves the overall scene
structure with substantially fewer initial anchors, while GBSP further improves
the local structural consistency and detail preservation of the complete model.

The scene-wise results in Table~\ref{tab:ablation_bungee_scene} further confirm
the consistent effectiveness of the two components. GBAI reduces the initial
anchor count by 35.5\%--45.8\% across all six scenes, whereas GBSP further
decreases the final anchor count and storage in every scene. The most notable
improvement is observed on \textit{Pompidou}, where the complete model reduces
the final anchor count from 621.8K to 480.8K and storage from 188 MB to 145 MB,
while improving all three rendering metrics. Although the quality trade-off
varies across scenes, the complete model achieves the best overall balance
between representation compactness and rendering quality.

\begin{table*}[!htbp]
\centering
\caption{
Scene-wise ablation results on BungeeNeRF~\cite{xiangli2022bungeenerf}.
SfM point and anchor counts are reported in thousands, and storage is reported in MB.
}
\label{tab:ablation_bungee_scene}
\setlength{\tabcolsep}{3.2pt}
\renewcommand{\arraystretch}{1.12}
\resizebox{\textwidth}{!}{
\begin{tabular}{l|l|cccc|ccc}
\toprule
Scene & Method 
& SfM Points (K)
& Init. Anchors (K)$\downarrow$
& Final Anchors (K)$\downarrow$
& Storage$\downarrow$
& PSNR$\uparrow$
& SSIM$\uparrow$
& LPIPS$\downarrow$ \\
\midrule

\multirow{3}{*}{Rome}
& Scaffold-GS 
& 123.9K & 120.8K & 564.7K & 171 
& 22.962 & 0.835 & 0.200 \\
& + GBAI
& 123.9K & 66.4K & 557.0K & 168 
& 23.538 & 0.839 & 0.211 \\
& + GBAI + GBSP 
& 123.9K & 66.4K & 528.9K & 160 
& 23.276 & 0.838 & 0.208 \\
\midrule

\multirow{3}{*}{Hollywood}
& Scaffold-GS 
& 110.3K & 106.4K & 452.3K & 137 
& 23.718 & 0.767 & 0.258 \\
& + GBAI
& 110.3K & 65.6K & 460.2K & 139 
& 23.342 & 0.752 & 0.293 \\
& + GBAI + GBSP 
& 110.3K & 65.6K & 428.0K & 130 
& 23.283 & 0.756 & 0.281 \\
\midrule

\multirow{3}{*}{Quebec}
& Scaffold-GS 
& 117.4K & 113.5K & 480.2K & 145 
& 28.235 & 0.921 & 0.153 \\
& + GBAI
& 117.4K & 66.2K & 464.5K & 141 
& 28.337 & 0.918 & 0.165 \\
& + GBAI + GBSP 
& 117.4K & 66.2K & 441.6K & 134 
& 28.252 & 0.921 & 0.153 \\
\midrule

\multirow{3}{*}{Bilbao}
& Scaffold-GS 
& 87.5K & 84.3K & 659.2K & 199 
& 27.981 & 0.892 & 0.156 \\
& + GBAI
& 87.5K & 54.4K & 542.6K & 164 
& 27.791 & 0.887 & 0.166 \\
& + GBAI + GBSP 
& 87.5K & 54.4K & 522.5K & 158 
& 27.921 & 0.890 & 0.165 \\
\midrule

\multirow{3}{*}{Amsterdam}
& Scaffold-GS 
& 130.4K & 124.8K & 699.9K & 211 
& 26.736 & 0.896 & 0.144 \\
& + GBAI
& 130.4K & 67.7K & 684.2K & 207 
& 26.640 & 0.886 & 0.164 \\
& + GBAI + GBSP 
& 130.4K & 67.7K & 659.2K & 199 
& 26.738 & 0.918 & 0.164 \\
\midrule

\multirow{3}{*}{Pompidou}
& Scaffold-GS 
& 135.2K & 130.9K & 621.8K & 188 
& 19.715 & 0.709 & 0.301 \\
& + GBAI
& 135.2K & 71.0K & 576.1K & 174 
& 20.121 & 0.722 & 0.299 \\
& + GBAI+ GBSP 
& 135.2K & 71.0K & 480.8K & 145 
& 20.205 & 0.728 & 0.294 \\
\midrule

\multirow{3}{*}{Average}
& Scaffold-GS 
& 117.4K & 113.5K & 579.7K & 175.2 
& 24.891 & 0.837 & 0.202 \\
& + GBAI
& 117.4K & 65.2K & 547.4K & 165.5 
& 24.962 & 0.834 & 0.216 \\
& + GBAI + GBSP 
& 117.4K & 65.2K & 510.2K & 154.3 
& 24.946 & 0.842 & 0.211 \\
\bottomrule
\end{tabular}
}
\end{table*}

The scene-wise results in Table~\ref{tab:ablation_bungee_scene} show that granular ball centers consistently reduce the number of initial support units across all six scenes. Adding the scale prior further decreases the final model size in every scene, with particularly notable reductions on Pompidou. Although rendering-quality changes vary across scenes, the complete model achieves the best overall balance, reducing the average final support-unit count from 579.7K to 510.2K and storage from 175.2 MB to 154.3 MB while maintaining comparable average PSNR and SSIM.

\subsection{Anchor Growth and Computational Cost Analysis}
\label{sec:anchor_growth_efficiency}
To examine whether the compact initialization of 3DGBGS can be maintained
during subsequent optimization, we vary the stopping iteration of dynamic
anchor growth on the \textit{Bonsai} and \textit{Bicycle} scenes. We report
the final anchor count, the measured GFLOPs, and rendering quality under
different anchor-growth schedules.

As shown in Table~\ref{tab:anchor_growth}, extending the anchor-growth period
generally increases both the final anchor count and the measured computational
cost for Scaffold-GS and 3DGBGS. Nevertheless, under the same growth schedule,
3DGBGS consistently maintains fewer final anchors and lower GFLOPs than
Scaffold-GS. Under the full-growth setting, 3DGBGS reduces the final anchor
count and GFLOPs by 15.46\% and 29.58\%, respectively, on
\textit{Bonsai}, while the PSNR decreases by only 0.401~dB. On
\textit{Bicycle}, it reduces the two quantities by 4.40\% and 4.98\%,
respectively, with a PSNR decrease of only 0.087~dB.

These results indicate that the compact spatial representation established
by GBAI and GBSP is largely preserved during dynamic anchor growth, reducing
the computational cost of the counted network components while maintaining
comparable rendering quality. It should be noted that the reported GFLOPs do
not include the computation of the custom CUDA rasterization kernels and
therefore do not represent the complete end-to-end rendering cost.

\begin{table*}[!t]
\centering
\caption{
Anchor growth and computational cost on \textit{Bonsai} and
\textit{Bicycle} from Mip-NeRF 360~\cite{barron2022mip}.
All models are trained for 30K iterations. ``5K'' and ``10K'' indicate that
dynamic anchor growth is enabled only during the first 5K and 10K iterations,
respectively. The reported GFLOPs exclude custom CUDA rasterization kernels.
}
\label{tab:anchor_growth}
\setlength{\tabcolsep}{4.5pt}
\renewcommand{\arraystretch}{1.08}
\resizebox{\textwidth}{!}{
\begin{tabular}{cclccc}
\toprule
Scene
& Growth
& Method
& Final Anchors (K)$\downarrow$
& GFLOPs$\downarrow$
& PSNR$\uparrow$ / SSIM$\uparrow$ / LPIPS$\downarrow$ \\
\midrule

\multirow{6}{*}{Bonsai}
& \multirow{2}{*}{5K}
& Scaffold-GS
& 288.90 & 5.375
& 32.539 / 0.946 / 0.184 \\
&
& 3DGBGS (Ours)
& \textbf{233.06} & \textbf{3.819}
& 32.128 / 0.944 / 0.187 \\
\cmidrule(lr){2-6}

& \multirow{2}{*}{10K}
& Scaffold-GS
& 360.42 & 5.584
& 32.698 / 0.948 / 0.180 \\
&
& 3DGBGS (Ours)
& \textbf{305.90} & \textbf{4.019}
& 32.382 / 0.945 / 0.183 \\
\cmidrule(lr){2-6}

& \multirow{2}{*}{Full}
& Scaffold-GS
& 416.09 & 5.756
& 32.834 / 0.948 / 0.179 \\
&
& 3DGBGS (Ours)
& \textbf{351.75} & \textbf{4.053}
& 32.433 / 0.946 / 0.182 \\
\midrule

\multirow{6}{*}{Bicycle}
& \multirow{2}{*}{5K}
& Scaffold-GS
& 338.20 & 1.880
& 24.765 / 0.720 / 0.293 \\
&
& 3DGBGS (Ours)
& \textbf{308.35} & \textbf{1.672}
& 24.693 / 0.714 / 0.300 \\
\cmidrule(lr){2-6}

& \multirow{2}{*}{10K}
& Scaffold-GS
& 655.20 & 4.721
& 25.059 / 0.740 / 0.265 \\
&
& 3DGBGS (Ours)
& \textbf{615.24} & \textbf{4.469}
& 25.019 / 0.736 / 0.271 \\
\cmidrule(lr){2-6}

& \multirow{2}{*}{Full}
& Scaffold-GS
& 912.30 & 7.187
& 25.143 / 0.746 / 0.257 \\
&
& 3DGBGS (Ours)
& \textbf{872.15} & \textbf{6.829}
& 25.056 / 0.742 / 0.262 \\
\bottomrule
\end{tabular}
}
\end{table*}

\subsection{Hyperparameter Analysis}
\label{sec:hyperparameter}

We investigate the influence of the maximum Granular Ball partition depth
$D_{\max}$ on the two indoor scenes of Deep Blending. A larger $D_{\max}$
allows spatially complex regions to be partitioned more finely, but may also
increase the number of initial anchors. We therefore evaluate its effects on
both representation compactness and rendering quality. The numbers of input
SfM points remain fixed at 80.861K for \textit{DrJohnson} and 37.005K for
\textit{Playroom} under all settings and are thus omitted from the table.
As shown in Table~\ref{tab:gb_depth}, increasing $D_{\max}$ from 10 to 18
raises the average number of initial anchors from 1.02K to 34.14K. Despite
this more than 33-fold variation, the final anchor count remains within a
relatively narrow range of 151.09K--163.68K and does not vary monotonically
with the partition depth. In particular, $D_{\max}=14$ yields the smallest
final anchor count, although it does not produce the fewest initial anchors.
These results indicate that the final representation size is not directly
determined by the initialization size, since dynamic anchor growing and
pruning can compensate for different Granular Ball partition granularities.

Rendering quality generally improves as $D_{\max}$ increases from 10 to 16
and then becomes stable. The maximum PSNR difference among
$D_{\max}=14$, 16, and 18 is only 0.143~dB, indicating that the rendering
performance largely saturates once sufficient partitioning capacity is
provided. Although $D_{\max}=16$ achieves the best average PSNR and LPIPS,
its differences from $D_{\max}=18$ are only 0.023~dB and 0.001,
respectively, while both settings obtain the same average SSIM of 0.905.

We therefore set $D_{\max}=18$ as a permissive upper bound in all experiments.
Since the adaptive stopping criteria terminate the partitioning of sufficiently
compact and appearance-consistent granular balls, increasing the maximum depth
does not force every region to undergo additional splitting. This is reflected
by the nearly identical initial anchor counts at depths 16 and 18. Consequently,
$D_{\max}=18$ preserves sufficient partitioning capacity for structurally
complex regions without substantially increasing the initialization size.

\begin{table*}[!t]
\centering
\caption{
Effect of the maximum Granular Ball partition depth on Deep
Blending~\cite{hedman2018deep}. Initial and final anchor counts are averaged
over \textit{DrJohnson} and \textit{Playroom} and reported in thousands.
}
\label{tab:gb_depth}
\setlength{\tabcolsep}{3.5pt}
\renewcommand{\arraystretch}{1.10}
\resizebox{\textwidth}{!}{
\begin{tabular}{c|cc|ccc|ccc|ccc}
\toprule
\multirow{2}{*}{$D_{\max}$}
& \multicolumn{2}{c|}{Avg. Anchors (K)}
& \multicolumn{3}{c|}{DrJohnson}
& \multicolumn{3}{c|}{Playroom}
& \multicolumn{3}{c}{Average} \\
\cmidrule(lr){2-3}
\cmidrule(lr){4-6}
\cmidrule(lr){7-9}
\cmidrule(lr){10-12}
& Init. & Final
& PSNR$\uparrow$ & SSIM$\uparrow$ & LPIPS$\downarrow$
& PSNR$\uparrow$ & SSIM$\uparrow$ & LPIPS$\downarrow$
& PSNR$\uparrow$ & SSIM$\uparrow$ & LPIPS$\downarrow$ \\
\midrule
10
& 1.02 & 163.68
& 28.981 & 0.889 & 0.284
& 30.446 & 0.903 & 0.261
& 29.713 & 0.896 & 0.273 \\
12
& 4.10 & 161.78
& 29.152 & 0.895 & 0.275
& 30.629 & 0.904 & 0.263
& 29.890 & 0.899 & 0.269 \\
14
& 16.37 & 151.09
& 29.294 & 0.899 & 0.268
& 30.713 & 0.905 & 0.259
& 30.003 & 0.902 & 0.264 \\
16
& 34.13 & 157.19
& 29.596 & 0.904 & 0.256
& 30.696 & 0.906 & 0.257
& 30.146 & 0.905 & 0.256 \\
18
& 34.14 & 160.65
& 29.513 & 0.903 & 0.257
& 30.732 & 0.906 & 0.256
& 30.123 & 0.905 & 0.257 \\
\bottomrule
\end{tabular}
}
\end{table*}

\section{Conclusion}
\label{sec:conclusion}

In this work, we presented 3DGBGS, a compact anchor-based Gaussian splatting
framework that adaptively organizes spatially non-uniform SfM point clouds
using 3D granular balls. The name 3DGBGS reflects its twofold connection to
Granular-Ball Computing: the coarse-to-fine and on-demand generation of
multi-scale Gaussian ellipsoids in 3DGS is highly consistent with the GBC
paradigm, while our method further introduces granular-ball partitioning to
optimize anchor construction. GBAI employs granular-ball centers to construct
a compact anchor layout, while GBSP incorporates granular-ball radii as local
scale priors. Experiments on 19 scenes from four real-world benchmarks show
that 3DGBGS reduces the numbers of initial and final anchors by 37.1\% and
10.0\%, respectively, and model storage by 9.8\% on average, while maintaining
rendering quality comparable to Scaffold-GS. These results demonstrate the
effectiveness of adaptive granular-ball organization for compact novel-view
synthesis. Future work will explore anisotropic granular representations and
learnable partitioning strategies to further improve the compactness--quality
trade-off in structurally complex scenes.

\bibliography{ref}
\bibliographystyle{IEEEtran}





\end{document}